\documentclass{bmvc2k}

\title{Representation-Aggregation Networks for Segmentation of Multi-Gigapixel Histology Images}

\addauthor{\textbf{$^*$Abhinav Agarwalla}}{agarwallaabhinav@gmail.com}{1}
\addauthor{\textbf{$^*$Muhammad Shaban}}{m.shaban@warwick.ac.uk}{2}
\addauthor{Nasir M. Rajpoot }{n.m.rajpoot@warwick.ac.uk}{2}

\addinstitution{
 Department of Mathematics,\\
 Indian Institute of Technology, Kharagpur\\
 West Bengal 721302, India
}
\addinstitution{
 Department of Computer Science,\\ 
 University of Warwick, \\
 Coventry, CV4 7AL, UK
}

\runninghead{Agarwalla, Shaban, Rajpoot}{Representation-Aggregation Networks}

\def\etal{\emph{et al}\bmvaOneDot}

\begin{document}

\maketitle

\begin{abstract}
Convolutional Neural Network (CNN) models have become the state-of-the-art for most computer vision tasks with natural images. However, these are not best suited for multi-gigapixel resolution Whole Slide Images (WSIs) of histology slides due to large size of these images. Current approaches construct smaller patches from WSIs which results in the loss of contextual information. We propose to capture the spatial context using novel Representation-Aggregation Network (RAN) for segmentation purposes, wherein the first network learns patch-level representation and the second network aggregates context from a grid of neighbouring patches. We can use any CNN for representation learning, and can utilize CNN or 2D-Long Short Term Memory (2D-LSTM) for context-aggregation. Our method significantly outperformed conventional patch-based CNN approaches on segmentation of tumour in WSIs of breast cancer tissue sections.
\end{abstract}

%-------------------------------------------------------------------------
\section{Introduction}
\label{sec:intro}
\begin{figure}
\centering
\includegraphics[height=6cm]{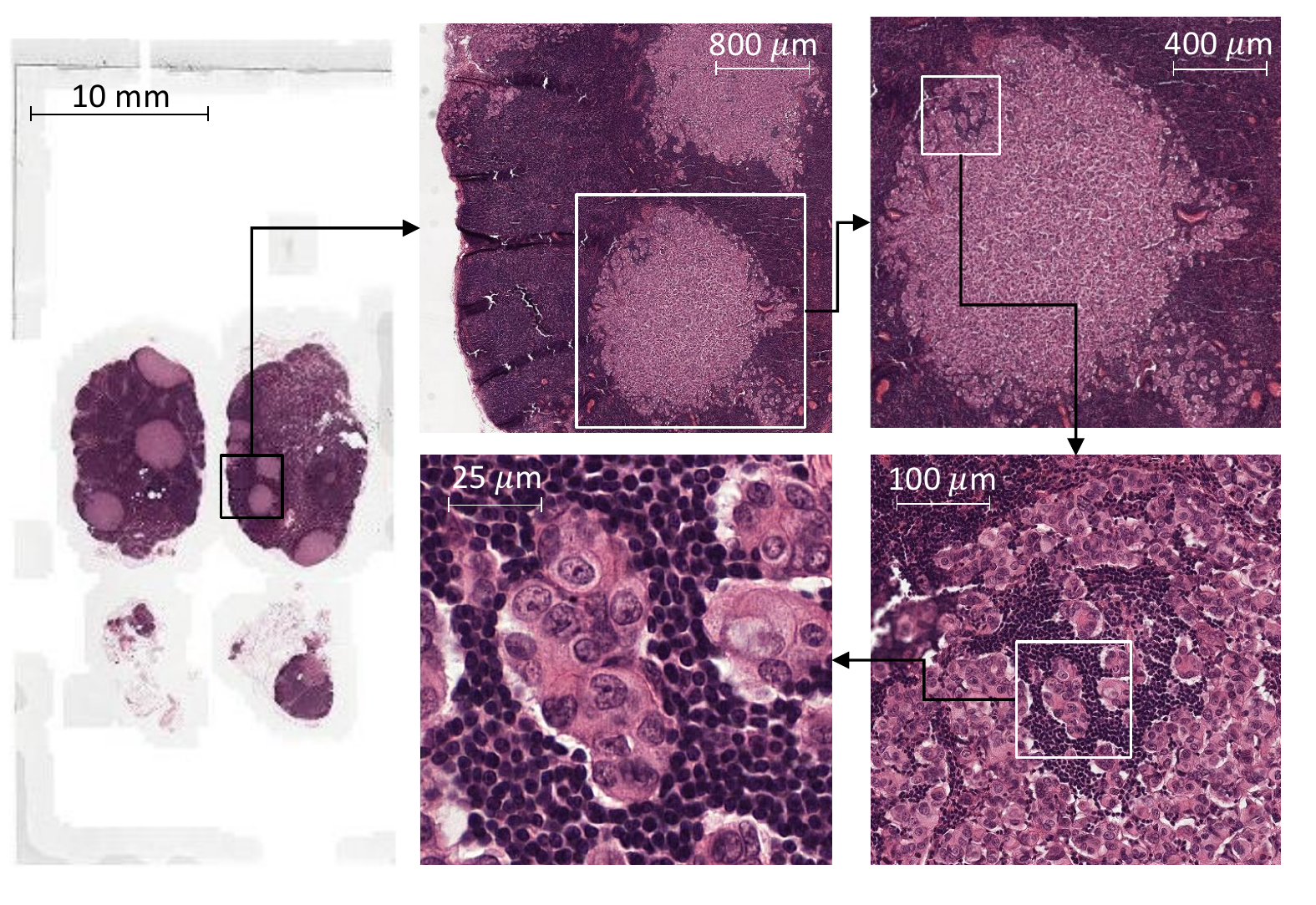}
\caption{A whole slide image and multi-scale visualization of a sub region.}
\label{fig:wsi}
\end{figure}

Recent technological developments in digital imaging solutions have led to wide-spread adoption of whole slide imaging (WSI) in digital pathology which offers unique opportunities to quantify and improve cancer treatment procedures. Stained tissue slides are digitally scanned to produce digital slides~\cite{wsi} at different resolutions till $40\times$ as shown in Figure~\ref{fig:wsi}. These digital slides result in an explosion of data which leads to new avenues of research for computer vision, machine learning and deep learning communities. Moreover, these multi-gigapixel histopathological WSIs can be excellently absorbed by data hungry deep learning methods to tackle digital pathology problems.

Convolutional Neural Network (CNN) models have significantly improved the state-of-the-art in many natural image based problems such as visual object detection and recognition~\cite{vggnet,frcnn} and scene labelling~\cite{scene_labelling}. However, classification of WSIs through a CNN raises serious challenges due to multi-gigapixel nature of images. Feeding the complete WSI or resizing WSI either leads to computationally unfeasible methods or loss of crucial cell level features essential for segmentation. This results in processing WSIs which are typically $200$K $\times100$K pixels in size in a patch-by-patch manner. Since patch based approaches face difficulties in handling images larger than a few thousand pixels, therefore using larger patches to capture maximum context is not a solution. A huge difference between patch size and WSI size results in loss of global context information which is extremely important for many tumour classification tasks~\cite{capc}.
% and video classifiction~\cite{lsvideo}.
% \begin{figure}
% \centering
% \includegraphics[height=3cm]{./low_res.pdf}
% \caption{Visualization of cell level features in a patch from (a) orginal WSI and (b) downsampled WSI.}
% \label{fig:lowres}
% \end{figure}

We propose Representation-Aggregation Networks (RANs) to efficiently model spatial context in multi-gigapixel histology images. RANs employ a representation learning network as a CNN which encodes the appearance and structure of a patch as a high dimensional feature vector. This network can be any state-of-the-art network such as AlexNet~\cite{alexnet}, GoogLeNet~\cite{googleNet}, VGGNet~\cite{vggnet} or ResNet~\cite{resnet}. A 2D-grid of features is generated by packing feature vectors for neighbouring patches in the WSI as encoded by the representation learning network. The first variant of context-aggregation network (RAN-CNN) in RAN utilizes a CNN with only convolutional and dropout layers. RAN-CNN takes input as a 2D-grid and outputs a tumour probability for each cell in the 2D-grid.

\begin{figure}
\centering
\includegraphics[height=6cm]{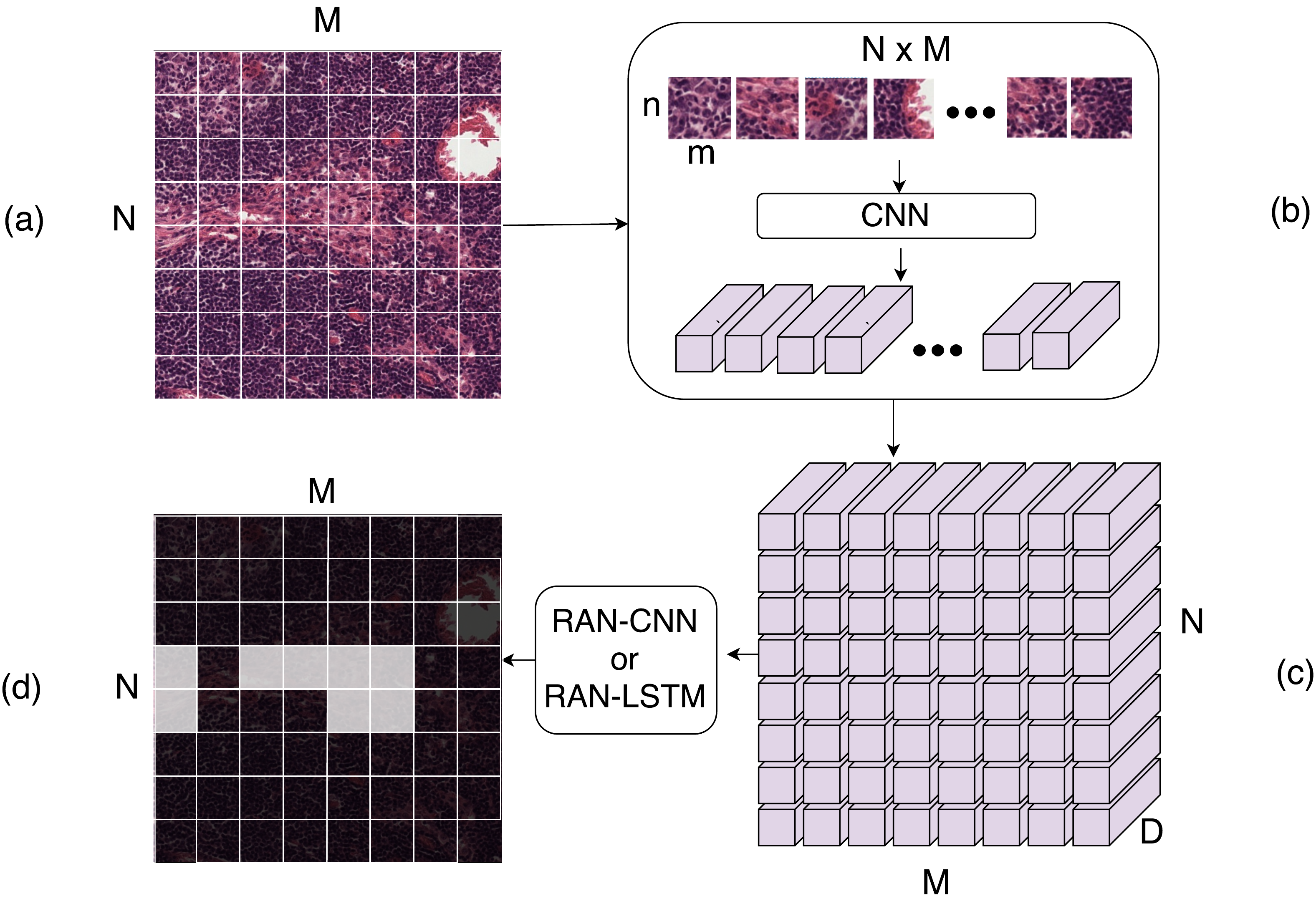}
\caption{(a) A large region from WSI which consists of NM patches. (b) A CNN (e.g. AlexNet, GoogLeNet, etc) encodes each patch independently into high dimensional features. (c) Rearranged features into 2D-grid format. (d) Overlay of prediction from RANs on the input. Both light and dark patches represent different classes.}
\label{fig:workflow}
\end{figure}

% In first RAN, we use a context based convolutional neural network (RAN-CNN) that consist of set of convolutuional and dropout layers without any max pooling layer. It the take the feature-cube as input and predict the label of each cell of the 2D-grid using the context of neighbouring cells. The aim of RAN-CNN is just to learn the context information because the input features already have good representation of cell appearance. Therefore, RAN-CNN is only consist of few convolutional layer and take less time in training. 

Recurrent Neural Networks (RNNs) along with their variants Long Short Term Memory (LSTM)~\cite{lstm} and Gated Recurrent Units (GRUs)~\cite{gru} have excelled at modelling sequences in challenging tasks like machine translation and speech recognition. We build the second variant of RAN (RAN-LSTM) by combining CNNs with 2D-LSTMs. RAN-LSTM captures the context information by treating WSIs as a two-dimensional sequence of patches. RAN-LSTM extends 2D-LSTMs for tumour segmentation task in multi-gigapixel histology images by using learned representations of neighbouring patches from represenation learning network as a context for tumour classification of a single patch. RAN-LSTM is constituted by four 2D-LSTMs running diagonally, one from each corner. Tumour predictions across all the dimensions are averaged together to get the final tumour classification. The complete workflow of the proposed architecture is shown in Figure~\ref{fig:workflow}.

We demonstrate the effectiveness of modelling context using RANs for tumour segmentation. RANs significantly outperform traditional methods on the dataset from Camelyon'16 challenge~\cite{camelyon16} on all metrics. Our main contributions can be summarized as follows:
\begin{itemize}
    \item We propose RANs as a generic architecture for context modelling in multi-gigapixel images.
    \item We utilize both CNNs and 2D-LSTMs for context-aggregation network.
    \item We show the effectiveness of the addition of context-aggregation network on top of a representation network for segmentation of tumour areas in multi-gigapixel histology images.
\end{itemize}

\section{Related Work}
With large memory storage and fast computational power available in modern machines, processing WSIs has become feasible. Recent studies have exploited WSIs for cell detection and classification~\cite{korsuk}, nuclei segmentation~\cite{nuclei_seg} and tumour segmentation~\cite{tumourSeg}. Both these approaches follow a patch based approach to process a WSI which significantly limits the available context information. Bejnordi et al.~\cite{capc} proposed a similar approach for breast tissue classification by using large input patches and stacking CNNs together. To deal with large input patches, the network is trained in two steps. On the other hand, RANs generalize the segmentation task through context-aggregation from encoded representations of a 2D-grid of small patches. RANs can incorporate CNNs, 2D-LSTMs or a combination of both for modelling spatial context in WSIs.

Multi-dimensional RNNs~\cite{mdrnn} have been employed to model sequences in both temporal and spatial dimensions. Recent approaches~\cite{reseg}~\cite{scene_lstm} model spatial sequences in an image to accomplish dense output for semantic segmentation tasks. Byeon \etal~\cite{scene_lstm} utilized four 2D-LSTMs running in each direction, whereas Visin \etal.~\cite{reseg} employed two bi-directional RNNs as two layers for up-down and left-right spatial modelling. The key difference between these two and our approach is that we try to model spatial context by aggregating multiple patches as a 2D-grid of patches instead of modelling spatial context within a single patch. Both ~\cite{scene_lstm} and ~\cite{reseg} model spatial context for natural images, whereas RANs can model much larger context in multi-gigapixel images.

\section{Representation-Aggregation Networks}
The proposed Representation-Aggregation Networks (RANs) have a two-network architecture wherein the first network learns patch-level representation, which is passed on to the second context-aggregation network. RAN is able to incorporate context from a large region by aggregating the learned features from the first network as 2D-grid of patches. RANs analyze a 2D-grid of patch-level features at once, and predict tumour probabilities for each cell in the grid by feeding representation of neighbouring cells as context.

The first network is essentially a representation learning network. It takes in input patches of size $n\times m\times3$ and yields a $D$-dimensional representation. One can use any state-of-the-art image classifier for this purpose. For our experiments, we train AlexNet~\cite{alexnet} on our dataset to classify patches as tumour or non-tumour. $D$-dimensional representations, denoted by $p_{t}$ are obtained by extracting features from an intermediate layer of a trained network. We experiment with various intermediate layers with later layers being more task specific. We discuss the proposed variants for context-aggregation network in the following subsections.

\subsection{RAN-CNN}
Convolutional Neural Networks are good at learning the spatial relations from the input. RAN-CNN is designed to capture spatial context from the neighbouring patches. It consists of five $3\times3$ convolutional layers. The first convolutional layer takes in 2D-grid feature as an input and subsequent layers operate on the output from the previous layer. One can control the context region by varying the convolutional filter size. Last three convolutional layers are followed by dropout layers to avoid overfitting.

\begin{align} 
\mathbf y_i = \mathbf f_{conv}(\mathbf p_t,\mathbf W_i) \circ \mathbf f_{a}(\cdot) \\
\mathbf y_j = \mathbf f_{conv}(\mathbf p_i,\mathbf W_j) \circ \mathbf f_{a}(\cdot)\circ\mathbf  f_{d}(\cdot) \\
\mathbf y = \mathbf f_{conv}(\mathbf p_j,\mathbf W) \circ \mathbf f_{a}(\cdot)
\end{align}where $f_{conv}$, $f_{a}$ and $f_{d}$ are the convolution, activation and dropout functions respectively; $W_i$, $W_j$ and $W$ are the trainable weights; the operator ($\circ$) provides the output of preceding function to the superseding function and operator ($\cdot$) represents the output of the preceding function; $y_{i}$ and $y_{j}$ are the outputs of $i^{th}$ and $j^{th}$ layers, $i \in {\{1,2,\dots,C\}}$, $j \in {\{1,2,\dots,D\}}$ and $C, D$ are the number of convolutional layers with and without dropout layers. $y$ represents the output of final prediction layer which maps the number of feature maps from $y_{d}^{th}$ layer to the total number of classes.

\subsection{RAN-LSTM}
\begin{figure}
\centering
\includegraphics[height=6cm]{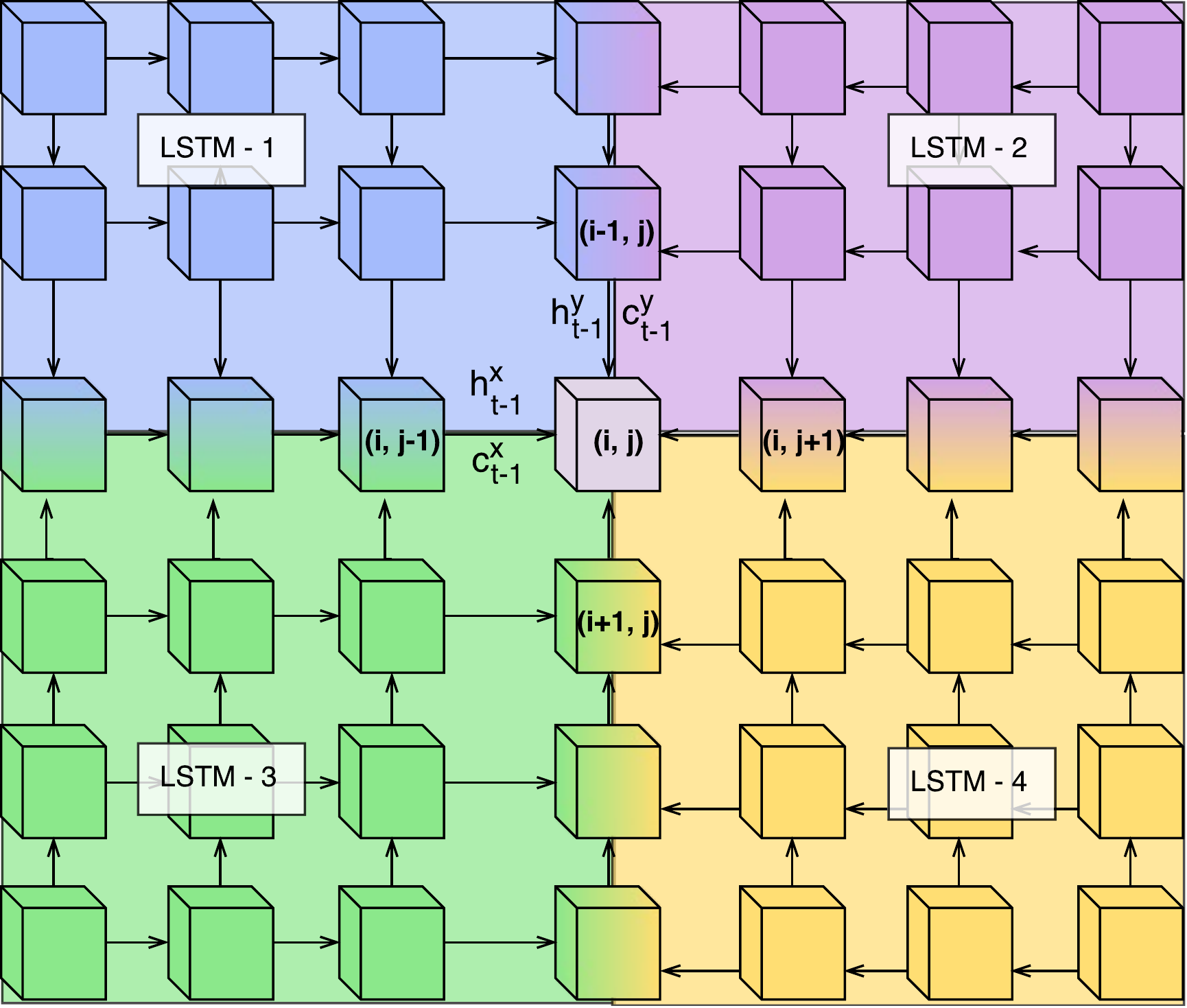}
\caption{Hidden states \{$h^x_{t-1}$, $h^y_{t-1}$\} and cell states \{$c^x_{t-1}$, $c^y_{t-1}$\} are used for prediction at $(i,j)$ by LSTM-1. Similarly, outputs from LSTM-{1,2,3,4} denoted by blue, pink, yellow and green are averaged for prediction at each cell $(i,j)$ by using context from adjacent patches.}
\label{fig:lstm}
\end{figure}

For modelling image sequences through standard 1D-LSTM, a sequence of $D$-dimensional representation is used as an input to the LSTM. On the other hand, 2D-LSTMs take two-dimensional inputs represented as a sequence of two $D$-dimensional vectors and generate either sparse or dense output predictions as required by the task. RAN-LSTM extends 2D-LSTM to model the context information along a 2D-grid of patches. Each 2D-LSTM unit $(i, j)$ has one input gate $(i_{t})$, two forget gates $(f^{x}_{t}, f^{y}_{t})$, two cell memory gates $(\tilde{c}^{x}_{t}, \tilde{c}^{y}_{t})$ and one output gate $(o_{t})$ for neighbouring patches in $x$ and $y$ direction respectively. The hidden states and cell states for current unit are denoted by $h_t$ and $c_t$ respectively. $h^{x}_{t-1}$ and $h^{y}_{t-1}$ denote hidden states for the neighbouring unit on left and top respectively. Similarly, $c^{x}_{t-1}$ and $c^{y}_{t-1}$ denote cell states for the neighbouring units. Unit $(i, j)$ is pairwise connected to its 4 neighbours i.e. $[(i-1, j), (i, j-1)], [(i-1, j), (i, j+1)], [(i+1, j), (i, j-1)], [(i+1, j), (i, j+1)]$ where each relation is exploited by an independent 2D-LSTM as shown in Figure \ref{fig:lstm}. These four 2D-LSTMs run in different directions, one from each corner to the diagonally opposite corner. Final predictions are obtained by aggregating results from 2D-LSTM from all directions. The governing equations for 2D-LSTM are given below where $p_{t}$, $W_{*}$, $U_{*}$, $b_{*}$ denote input vector and weights matrices for hidden
states, inputs and constants respectively. $\sigma$, $tanh$ and $\odot$ denote sigmoid activation, hyperbolic tangent activation function and dot product respectively. 2D-LSTM can be treated as a layer which accepts input of size N$\times$M$\times$D and outputs predictions of size N$\times$M$\times$D, where H indicates the hidden dimension of 2D-LSTM layer. Multiple 2D-LSTM layers can be stacked one after another to form RAN-LSTM just as convolutional layers for RAN-CNN.

\begin{align}
    	\mathbf i_t & = \sigma(\mathbf W_{i}\{\mathbf h^{x}_{t-1}, \mathbf h^{y}_{t-1}\} + \mathbf U_i p_t + \mathbf b_i) \\
		\{\mathbf f^{x}_t, \mathbf f^{y}_t\} & = \sigma(\mathbf W_{f}\{\mathbf h^{x}_{t-1}, \mathbf h^{y}_{t-1}\} + \mathbf U_f p_t + \mathbf b_f) \\
		\{\mathbf{\tilde{c}^{x}_t}, \mathbf{\tilde{c}^{y}_t}\} & = \tanh(\mathbf W_{c}\{\mathbf h^{x}_{t-1}, \mathbf h^{y}_{t-1}\} + \mathbf U_c p_t + \mathbf b_c) \\
		\mathbf o_t & = \sigma(\mathbf W_{o}\{\mathbf h^{x}_{t-1}, \mathbf h^{y}_{t-1}\} + \mathbf U_o p_t + \mathbf b_o) \\
		\mathbf c_t & =  \mathbf i_t \odot \mathbf{\tilde{c}_t} + \mathbf f^{x}_t \odot \mathbf c^{x}_{t-1}+ \mathbf f^{y}_t \odot \mathbf c^{y}_{t-1}\\
        \mathbf h_t & = \mathbf o_t \odot \mathbf c_t
\end{align}

Both variants are trained to minimize cross-entropy loss $L$ for 2D-grid as given below, where $y'_{i,j}$, $P(y_{i,j})$ denote the ground truth label and the predicted tumour probability respectively.
\begin{equation}
\label{eq:loss}
    L = \frac{1}{NM}\sum^N_{i = 1} \sum^M_{j = 1} y'_{i,j}\textnormal{log\ } P(y_{i,j})
\end{equation}

\begin{figure}[t]
\centering
\includegraphics[width=13cm]{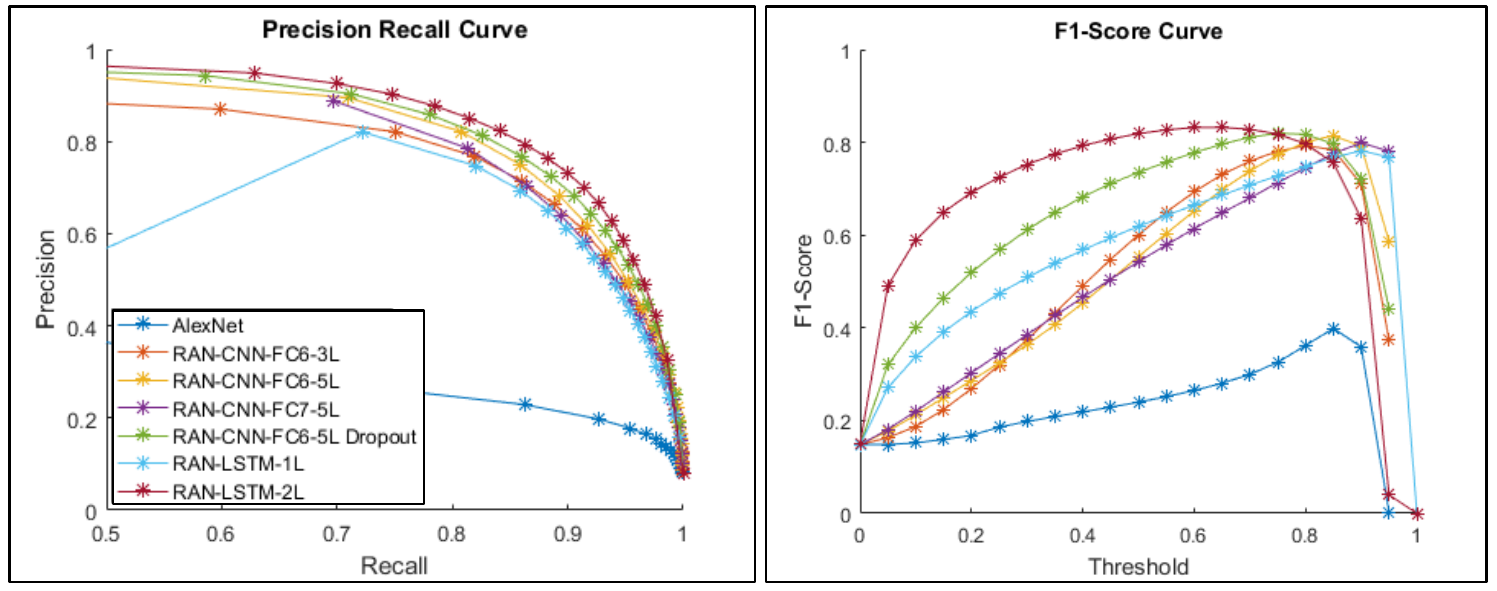}
\caption{Precision-Recall and F1-score curves of different experiments of RAN-CNN along with AlexNet and RAN-LSTM.}
\label{fig:prerecal}
\end{figure}

\section{Results and Discussion}

\textbf{The Dataset}. We evaluate our proposed method on the Camelyon'16 dataset~\cite{camelyon16} which consists of $110$ tumour and $160$ normal WSIs. For all WSIs, we extract the tissue region using a simple 2-layer Fully Convolution Network (FCN). We used $80\%$ WSIs for training and remaining for validation. Then, we randomly crop $188$K patches of size $224\times224$ to form the training set for patch-level network, out of which around $90$K are tumour patches. For training of context-aggregation networks, we extract a total of $190$K 2D-grids by aggregating $64$ $(N=8, M=8)$ patches together. For validation, $20$ complete WSI images are processed yielding a total of $99$K 2D-grids or $6$ million patches. Because of the relatively large size of training as well as validation dataset, we are fairly confident with the obtained scores and the generalization ability of our method.

\textbf{Model Specification}. For representation learning, we train AlexNet on the training set and experiment with $FC6$ and $FC7$ features as input to the context-aggregation networks. We fixed the context depth as $8$ and aggregated $64$ $(N=8, M=8)$ patches together as a 2D-grid to be processed by context-aggregation network, RAN-CNN or RAN-LSTM. We experimented with different network architectures of RAN-CNN to find a suitable one. First, we compared the impact of different number of convolutional layers and found that network with more convolutional layers performed better. Finally, we kept 5 convolution layers along with dropout for the last 3 convolutional layers, denoted by RAN-CNN-FC6-5L-D in Table \ref{tab:label}. After comparing the performance of $FC6$ and $FC7$ features, we decided to stick to $FC6$ features because of its superior performance. We experimented with the number of 2D-LSTM layers with 512 dimensional hidden state in each layers. Finally, we utilized two 2D-LSTM layers followed by a convolution layer to reduce the hidden state dimensions to the number of classes, which is 2 in our case.

\textbf{Training Details}. RAN-CNN model was trained using Adam optimizer with a batch size of 64. RAN-CNN converged after four epochs with total training time of 6 hours. For training RAN-LSTM, we used Adam optimizer with learning rate and decay rate as $0.0001$ and $0.5$ after every $2$ epochs respectively. The model is trained with a batch-size of $10$ for a total of $25$ epochs which took a total of 45 hours to train. All the codes were implemented in Tensorflow, and trained on a single NVIDIA GeForce GTX TitanX GPU. Out of $190$K training 2D-grids which is equivalent to $12$ million patches, only $6\%$ patches were tumorous. To tackle this class imbalance problem, we sample all 2D-grids that had at least one tumour patch along with the same number of non-tumour patches. This resulted in $28$K training 2D-grids for training the context-aggregation network in RANs.

\begin{figure}[t]
\centering
\includegraphics[width=12cm]{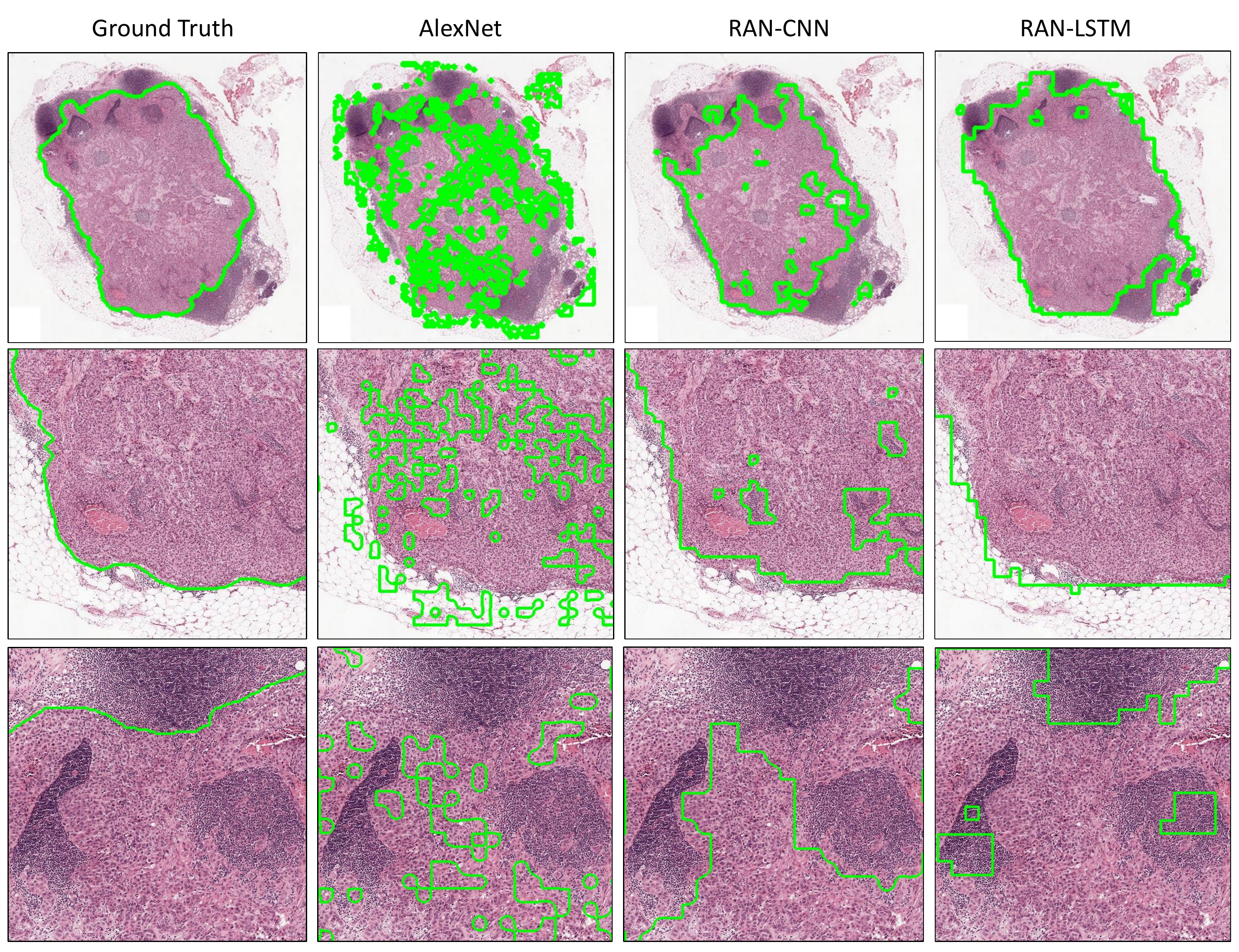}
\caption{Visual comparison of approaches along with ground truth where green color indicates the boundaries of a continuous tumour region.}
\label{fig:results}
\end{figure}

% \section{Results and Discussion}
We evaluate several variants of RAN using precision, recall and F1-score. We select F1-score as a metric for model performance instead of accuracy because of class imbalance in our data. Figure \ref{fig:prerecal} shows model performance through Precision-Recall curve and F1-scores at various thresholds. RANs lead to significant increase in F1-scores from 0.40 for AlexNet to 0.82, 0.83 for RAN-CNN and RAN-LSTM respectively. Since AlexNet classifies only a single patch at a time, the resultant predictions consist of several discontinuous blobs over the tumour region as shown in Figure \ref{fig:results}. This demonstrates the importance of context information while segmenting tumour region in multi-gigapixel histology images. The RAN-CNN and RAN-LSTM improve the prediction by incorporating the spatial context, and output smoother continuous regions. Thus, these are able to identify the global structure of the tumour region as opposed to AlexNet which only captures the local information from a single patch.

\begin{table}
\centering
\caption{Quantitative comparison of AlexNet, RAN-CNN and RAN-LSTM}
\label{tab:label}
\begin{tabular}{|l|l|l|l|}
\hline
Network & Precision     & Recall        & F1-Score      \\ \hline
AlexNet & 0.28          & 0.67          & 0.40          \\ \hline\hline
RAN-CNN-FC6-3L & 0.77      & 0.82          & 0.79          \\ \hline
RAN-CNN-FC6-5L & 0.82      & 0.81          & 0.81          \\ \hline
RAN-CNN-FC7-5L & 0.79      & 0.81          & 0.80          \\ \hline
RAN-CNN-FC6-5L-D    & 0.81          & \textbf{0.83} & 0.82          \\ \hline\hline
RAN-LSTM-1L & 0.74 & 0.82 & 0.78\\ \hline
RAN-LSTM-2L & \textbf{0.85} & 0.81          & \textbf{0.83} \\ \hline
\end{tabular}
\end{table}

Both variants of RAN achieve competitive results as summarized in Table~\ref{tab:label}. From the various different architectural variants of RAN-CNN using $FC6$ features with 5 convolution layers with dropout performs the best. RAN-LSTM with a single layer (RAN-LSTM-1L) is not able to perform well due to underfitting. A two layered RAN-LSTM (RAN-LSTM-2L) gives much better performance than RAN-LSTM-1L. We refer to RAN-CNN-FC6-5L-D as RAN-CNN and RAN-LSTM-2L as RAN-LSTM for convenience. RAN-CNN gives better recall of 0.83 as compared to 0.81 with RAN-LSTM but loses on precision with 0.81 and 0.85 for RAN-CNN and RAN-LSTM respectively. RAN-LSTM outperforms all the approaches yielding the best F1-score of 0.83. The superior performance of RAN-LSTMs may be attributed to its ability to capture global context of the complete 2D-grid at once, where as RAN-CNN generates output predictions largely from local context. From Figure \ref{fig:results}, we see that RAN-LSTM succeeds in modelling the entire tumour region as a single component whereas RAN-CNN has few discontinuities within the tumour region.

% \begin{figure}
% \centering
% \includegraphics[height=8cm]{./RAN-CNNF1.pdf}
% \caption{F1 score curve at different thresholds of different experiments of RAN-CNN along with AlexNet and RAN-LSTM}
% \label{fig:lowres}
% \end{figure}

\section{Conclusions}

Technical advances in digital scanning of tissue slides are posing unique challenges to the researchers in the area of digital pathology. These gigapixel tissue sides open the way for automated analysis of cancerous tissues by deep learning algorithms. We demonstrated how segmentation demands sophisticated deep learning approaches when dealing with multi-gigapixel histology images. We proposed Representation-Aggregation Network (RAN) as a generic network that can incorporate the context from the neighbouring patches to make global decisions on a task involving multi-gigapixel images. RANs can be easily modified by varying representation learning network and context-aggregation network with networks suited for a particular task. We evaluate the performance of RANs for the task of tumour segmentation where it outperforms standard CNN approaches by a large margin.

%future work: context with missing data semi-supervised setting; conv-lstm instead of 2D-LSTM because then the spatial features are not condensed within a patch
\bibliography{egbib}
\end{document}